\documentclass[twoside,11pt]{article}

\usepackage{jmlr2e}
\usepackage{algorithm}
\usepackage{algpseudocode}
\usepackage{amsmath}
\usepackage{hyperref}
\usepackage[utf8]{inputenc}
\usepackage{booktabs}

\DeclareMathOperator*{\argmax}{arg\,max}

\usepackage{graphicx}
\usepackage{mathtools}
\usepackage{footnote}
\usepackage{float}
\usepackage{xspace}
\usepackage{multirow}
\usepackage{wrapfig}
\usepackage{framed}
\usepackage{xcolor}

\newcommand{\field}[1]{\mathbb{#1}}

\newcommand{\fR}{\field{R}}

\newcommand{\calA}{{\mathcal{A}}}

\newcommand{\calD}{{\mathcal{D}}}

\newcommand{\calX}{{\mathcal{X}}}

\newcommand{\calN}{{\mathcal{N}}}

\jmlrheading{1}{}{}{}{}{}

\firstpageno{1}

\begin{document}

\title{A Deep Bayesian Bandits Approach for Anticancer Drug Screening : Exploration via Functional Prior}

\author{\name MingYu Lu  \email mingyulu@cs.washington.edu \\
\name Yifang Chen \email yfchen@cs.washington.edu \\
\name Su-In Lee  \email suinlee@cs.washington.edu \\
       \addr Department of Computer Science\\
       University of Washington\\
       Seattle, WA 98195-4322, USA}


\maketitle

\begin{abstract}
Identifying potential cancer treatments with machine learning holds great promise to precision oncology. 
However, anticancer drugs screening is a lengthy and cost-intensive process. To address this challenge, we formulate drug screening studies as a \emph{contextual bandit} problem, in which an algorithm selects anticancer therapeutics based on contextual information about cancer cell lines while adapting its treatment strategy to maximize treatment response. We propose using a novel \emph{deep Bayesian bandits} framework that uses functional prior to approximate posterior for drug response prediction based on genomic features and drug structure and leverage inherent information between actions. We empirically evaluate our method on three large-scale \emph{in-vitro} pharmacogenomic datasets and show that our approach outperforms several benchmarks in identifying the optimal treatment for a given cell line. 
\end{abstract}


\section{Introduction}
By using genomic signatures and identifying druggable targets, the landscape of precision oncology has progressed rapidly.\citep{10.1093/nar/gks1111, CCLE,IORIO2016740} One approach is to conduct anticancer therapeutics screening but this approach remains challenging as drug screening study is often an expensive, time-consuming process. Moreover, most of the studies rely on exhaustively experimenting all available drugs without considering cancer heterogeneity. To solve \emph{in-vitro} identification and treatment allocation in anticancer therapeutics screening more efficiently, a natural attempt is to cast it as a \emph{contextual bandit} problem, a well-known tool for personalization \citep{li2010contextual}. Contextual bandits framework enables effective personalized intervention by taking appropriate actions based on sample's characteristics instead of random or uniform treatment assignment. Recently, contextual bandits methods, especially Thompson Sampling \citep{thompsonSampling}, are being tailored to healthcare applications in a different domains, including mobile health \citep{IntelligentPooling} and risk stratification for multiple myeloma.\citep{DynamicRisk}

However, naively applying conventional bandit algorithms to drug screen studies is not feasible. First of all, the performance relies on accurate assumptions about the reward environment, which can be highly complex and non-linear. Second, finding a proper set of candidate models to map genomic features to drug sensitivity may also be difficult. Moreover, due to inherent heterogeneity across and within different cancer types and drug metabolism \citep{tumor_heterogeneity}, modeling uncertainty plays an important role in determining the right decision. Third, the exploration-exploitation dilemma adds another layer of complexity, meaning that more action-context pairs would be required to obtain a better approximation of different policies. Insufficient sample size or model of less expressive power would lead to a wrong decision. \citep{DynamicRisk} Finally, most conventional contextual bandit algorithms treat each action as an unrelated discrete point and fail to leverage the similarity and inherent structure between each action. For instance, in cancer trial, OTX015, a BET inhibitor (BETi) that exhibits anti-tumor activity in non-small cell and small cell lung cancer, shares a similar structure with JQ1 \citep{OTX015}. Under a conventional bandit setting, without knowing the inherent relationship between two similar drugs, unnecessary exploration would be conducted.

In this work, our contributions are as follows. First, we adapt a \emph{deep Bayesian bandits} approach in drug screening studies as it provides a natural solution to uncertainty estimation with high-dimensional data. Second, under this settings, to find an approach to approximate posterior distribution of drug response prediction function, we use `functional prior' \citep{hafner2019noise, sun2019functional, functionprior} and, following \citet{sun2019functional}, update functional posterior with variational inference through the Stein gradient estimator (SSGE)\citep{liu2019stein}. Third, to avoid unnecessary exploration, for functional posterior update, to select proper measurement data points,  we use a bootstrap procedure to sample from both perturbed genomics contexts and drug compound features. We empirically evaluated our method and compared it with other treatment allocation strategies using three large-scale drug sensitivity screens. Our method delivers the best performance against several benchmark algorithms.
\section{Background and Related Work}
\subsection{Contextual Bandits}
Contextual bandits \citep{li2010contextual} is a generalization of the multi-armed bandit in which the policy for choosing future actions is dependent on \emph{context} information at each step $t$. For example, in clinical trial, the context could be patients' symptoms or laboratory observations. 
It is characterized by a sequence of a context-reward pairs $\{x_t,r_t\}_{t=1,2,\ldots,T} \in \calX \times \fR^{|\calA|}$, where $\calX,\calA$ are some arbitrary context space and action space (e.g., a set of treatment choices), respectively. Given a candidate policy set $\Pi$, the contextual bandits problem aims to learn a policy $\pi: \calX \to \calA$, or a mapping from context to actions that maximizes cumulative rewards, or equivalently to minimize the total regret compared to the best policy in hindsight.



\subsection{Thompson Sampling \& Deep Bayesian Bandits}

Thompson Sampling is a Bayesian approach \citep{thompsonSampling,AnalysisofTS} that allows sampling from the posterior distribution, $P(\theta)$, over plausible problem instances such as rewards or model parameters. At each round $t$, it observes the context, $x_t$, and an arm $a$ is chosen according to the probability that it maximizes the expected reward. The posterior distribution is then updated after the result of an action is observed. Through the lens of Thompson Sampling, deep Bayesian bandits refer to tackling contextual bandits by parameterizing the action-value function and a prior distribution over $P_0(\theta)$ placed over the parameters. Although the weight posterior for neural networks is analytically intractable, it can be approximated by methods such as Dropout \citep{Dropout, JMLR:v15:srivastava14a} or Monte Carlo methods \citep{HMC}. \citet{riquelme2018deep} conduct a comprehensive analysis of different approximation methods in deep Bayesian bandit settings.

\section{Methods}
\subsection{Problem Setup}
We formulate treatment allocation in anti-cancer therapeutics as a Bayesian contextual bandit problem. That is, we have a context set $\calX \subset \fR^{d_1}$, defined as the individual genomics information with $d_1$-dimension; a discrete action set $\calA \subset \fR^{d_2}$, defined as a set of given drugs whose compounds can be expressed as a $d_2$-dimension feature vector; and a reward set $R \subset \fR^{|\calA|} $, where each reward vector is a possible collection of corresponding drug responses (pIC-50/Activity Area) for all drugs. Finally, we are given a policy set $\Pi: \calX \to \calA$, which is deep Bayesian neural net here. The protocol is as follows: 
\begin{itemize}
    \item At time $t$, we are given a context, $x_t \in \calX$, and a \textit{non-observable} reward vector $r_t \in R$, that are jointly drawn from some unknown $i.i.d$ distribution $\calD$. In our case, this scenario can be regarded as randomly sampled cancer cell line at time $t$ whose actual responses to various drugs are unknown.
    \item Then we play $a_t \in \calA$ based on previous history and observe \textit{only} the corresponding $r_t(a_t)$, for example, the drug response in our case, not in the whole $r_t$ vector.
\end{itemize}
By repeating this process for a total of $T$ times (or $T$ incoming samples), we can obtain the total cumulative reward
$\sum_{t=1}^T r_t(a_t).$
We measure the difference between our algorithm performance and the benchmark performance by the notion of \textit{cumulative regret} $R(T)$, defined as 
\begin{equation}\label{cumulative_regrets}
        R(T) = \max_{\pi^*}\sum_{t=1}^T r_t(\pi^*(x_t)) - \sum_{t=1}^T r_t(a_t).
\end{equation}



\subsection{Drug Response Prediction with Genomics and Drug Contexts}\label{3.2}

In precision oncology, treatment efficacy depends not only on patients' genomics features but also on therapeutics information which have shown to improve the predictive performance. \citep{predictingDrug_response, 10.1093/bioinformatics/btx806} 
Here, our solution is to treat both genomics features \emph{and} therapeutics context as a combined input into our neural network and show that the trained neural net can automatically capture inherent relationships between the two, genomics information (RNA expression) and drug features (MorganFingePrint); therefore, explore more efficiently. Specifically, on each round $t$ for each action $a$, we denote such concatenated feature vector as $x_{t} = ({x_g, x_d}) \in R^{d_1 + d_2}$, where $d_1$ and $d_2$ are the dimension of genomics and drug features, respectively. When the agent selects an action $a_d$ based on $x_g$, corresponding drug information $x_d$ will be provided for reward estimation.

\subsection{Functional Prior and Posterior Approximation for multi-modal Data}

Even though drug features can provide additional information in reward prediction, multimodal data of different data types complicates determining the prior distribution. The weight distribution 
would be highly variable and difficult to express with some commonly used prior distributions, such as the Gaussian distribution. \citep{hafner2019noise, functionprior} Therefore, we are not aiming to learn a posterior distribution of weights in weight space but a distribution of \emph{functional} posterior \citep{hafner2019noise, sun2019functional, functionprior} which is a function space that maps pharmacogenomics data to drug response. \citet{sun2019functional} introduce function space variance, analogous to weight space variational inference. 
\begin{equation}
    f(x) = g_{\phi}(x, \xi)
\end{equation}
Here $f(x)$ is a sampled function for some function $g_{\phi}$, a neural network with stochastic weights and stochastic inputs, and $\xi$, a random noise vector, corresponding to randomness in \emph{functional} space. For any given dataset $D$, the functional variance inference maximizes functional ELBO (fELBO) defined as:
\begin{equation}
    \mathcal{L}(q) = \mathbb{E}_q [\log(p(D|f))] - \sup_{n \in N, x \in \mathcal{X}^n} KL[q(f^X) || p(f^X)].
\end{equation}
As there is no analytical form for the functional KL divergence, following \citet{sun2019functional}, we also utilize the spectral Stein gradient estimator (SSGE) to approximate log density derivatives. SSGE is proposed by \citet{liu2019stein} is an estimator that only requires measurement samples from the targeted implicit distribution i.e., distributions without tractable densities, functional space for drug response prediction in out case.



\subsection{Sampling Measurement Sets for Cancer Treatment Exploration}

To estimate KL divergence properly with SSGE, we need unbiased samples from the implicit distribution of genomic contexts and drug features. Typically, one can sample from the training set where noise is injected into the data or from region of testing data. \citep{hafner2019noise}. Here we use a combination of two approaches \citep{functionprior}: to encourage exploration, we use sample not only from history $D_t$ but also from random drug feature $x_{a'}$ in the action space that we are interested in exploring. To address this problem, we use a simple bootstrapped method \citep{osband2015bootstrapped} to sample measurement sets. Accordingly, we sample a new history set $H$ from both training history $D_t$  and $\Tilde{D}_t$ with perturbed action-context-concatenated features $\Tilde{x}_{t}$. 
\begin{equation}
    \Tilde{D}_t = (\Tilde{x}_1, \dots \Tilde{x}_t); \Tilde{x}_t = (x_g, x_{\Tilde{d}}); \; H_t \sim D_t \cup \Tilde{D}_t.
\end{equation}
Therefore, with $H$ as our measurement set, $f$ELBO can be re-written as 
\begin{equation}
    \mathcal{L}_{H}(q) = \mathbb{E}_q[\log(p(D|f))] - \sup_{n \in N, x \in \mathcal{X}^n} KL\left[q(f^H)|| p(f^H)\right].
\end{equation}
With SSGE and proper measurement samples, we can approximate the entropy of both $q$ and $p$, which are both from implicit distribution; then, we can sample from the implicit posterior of drug response prediction for different compounds in bandits settings.  
\begin{algorithm}[h] 
\small
\caption{Functional Posterior Update}
\begin{algorithmic}[1]
    \State \textbf{Universal parameters} : $KL$ weight $\lambda$
    \State \textbf{Input:} $D_t$, functional variational posterior $g(\cdot|\phi)$
    \State  $\Tilde{D}_t \sim D_t$;\Comment{Sample $\Tilde{D}_t$ from $D$ with perturbed genomics and action contexts.}
    \State $f = g([\Tilde{D}_t, D_t],\mathcal{E}|\phi)$ \Comment{Sample a new function $f(.)$ with noise vector $\mathcal{E}$.} 
    \State  $\Delta_1 = \frac{1}{|D_t|}\nabla\phi \log_p(y|f_i)$ \Comment{Compute reconstruction loss.}
    \State $\Delta_2 = SSGE(\Tilde{D}_t, D_t)$ \Comment{Estimate KL divergence from implicit posterior.}
    \State $\phi \leftarrow \textnormal{optimize}(\phi, \Delta_1 - \lambda\Delta_2)$
    \State \textbf{Return } functional variational posterior $g(\cdot|\phi)$
\end{algorithmic}
\end{algorithm} 
\vspace*{-6mm}
\section{Experimental Design and Data}
\subsection{Simulated Drug Screen Design}
In this work, we simulate anticancer treatment trials and evaluate our algorithms in three large-scale cancer cell line screenings, including Genomics in Drug Sensitivity in Cancer (GDSC-1 and GDSC-2) \citep{10.1093/nar/gks1111, IORIO2016740} and the Cancer Cell Line Encyclopedia (CCLE) \citep{CCLE}. For cancer cell lines, we used RNA expressions and performed principal component analysis (PCA) to reduce the RNA expression dimension to 500. For drug context, we converted the drug SMILE string to Morgan fingerprint with 256 bits. \citep{MorganFingerPrint} For drug-dose response, in GDSC, we adapted negative log of half-maximal inhibitory concentration, $pIC50$. In CCLE, since some IC50 measurements are missing, we used activity area (AA) instead. All cancer cell lines and drug responses were scaled into the range of [0,1]. We run the bandits/simulated trials for 5000 rounds.


\subsection{Evaluation Metrics and Benchmark Algorithms}\label{baselines}
As described in Equation \ref{cumulative_regrets}, we reported cumulative regrets, the reward difference between competing algorithm, $\pi$, and an oracle, $\pi^*$, that would have access to the (unknown) treatment effect functions. We test our algorithm against competing algorithms (\textbf{Appendix A.4}) of different posterior approximations that demonstrated promising results. \citep{riquelme2018deep}  
\vspace*{-3mm}
\section{Results}\label{Results}
\subsection{Effect of Different Measurement Sets on the fBNN posterior}

\begin{figure}[h]
\centering
\begin{minipage}{.5\textwidth}
    \centering
    \includegraphics[scale = 0.24]{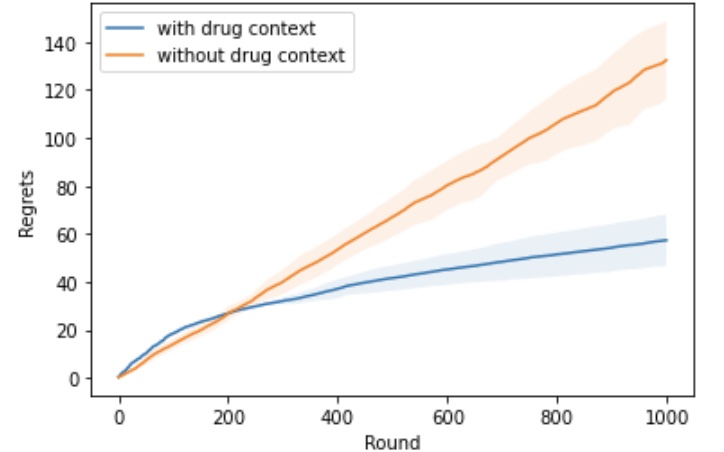}
\end{minipage}%
\begin{minipage}{.5\textwidth}
    \centering
    \includegraphics[scale = 0.24]{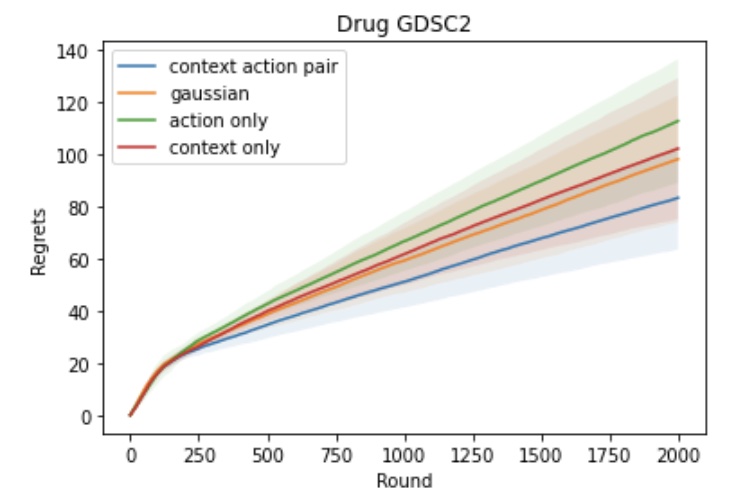}
\end{minipage}%
\caption{Cumulative regrets between (a) with and without drug context in the GDSC dataset, and (b) Functional Bayesian Neural Network with different measurement set samples. The confidential interval is generated through 20 trials.}
\label{fig:action_perturb}
\end{figure}

Here we show the effectiveness of using drug compound features as input contexts and adding action perturbation. From Figure \ref{fig:action_perturb}, we show that our strategy gives much lower cumulative regret compared to ones without them. To demonstrate that context-action pair perturbation as proxy of functional prior improves uncertainty estimation for drug response prediction, we compared the performance of models updated with different samples for SSGE, including (i) perturbing only genomics context, $x_g' = x_g + \epsilon; \epsilon \sim \calN(0,1)$ (ii) perturbing only action (drug) context $x_d' \in R^{d_2}$ and (iii) random sampling from multivariate Gaussian distribution, $x_g' \in R^{d_1}; x_g \sim \calN(0, 1)$ and random selected drugs from the action set, $x_d' \sim U(R^{d_2})$. Our sampling method, compared to the other proxy samples as in Figure 1, improves the quality of approximation of the desired prior and achieves lower regret. 

\subsection{Molecular Therapeutics Recommendation in Cancer Treatment }
Now we empirically evaluate our functional Bayesian neural bandit with benchmark algorithms in deep bayesian bandits in selecting drug compounds that would maximize drug response. Selecting optimal treatment requires not only learning drug response function but also uncertainty estimation for exploration. As shown in Table \ref{tab:regrets_summary}, models that can only optimize both, for example, NeuralGreedy and BBB performs poorly. \citep{riquelme2018deep}
Bootstrap and ParameterNoise seem to provide better posterior approximations for drug response prediction but suffer from under exploration, sometimes committing to the suboptimal drug in the experiments \citep{riquelme2018deep, sun2019functional}. Our method, on the other hand, by using genomics-action pair as proxy for functional prior, provides a better exploration strategy. 

\vspace{-3.5em}
\begin{table}[h!]
    \scriptsize
    \caption{Cumulative regret distribution at the final step in 3 drug screen studies. We report the mean and standard error of the mean over 20 trials. }
    \begin{center}
    \begin{tabular}{llll}
      \toprule
         \textbf{Algorithm}&  \textbf{GDSC-1} & \textbf{GDSC-2} & \textbf{CCLE}  \\
         \midrule
        Uniform & 2011 $\pm$ 10.2 & 995 $\pm$ 6.7 & 2509 $\pm$ 12.3\\
        NeuralGreedy & 531.4 $\pm$ 194.3 &  262.2 $\pm$ 152&1361.4  $\pm$ 361.7\\
        BayesByBackprop(BBB) & 1500.2 $\pm$396.4 & 497.3 $\pm$ 187.3 &  1682.3 $\pm$ 390.0\\
        DropOut & 508.3$\pm$ 33.4 &378.2 $\pm$ 84.0 & 1176.2 $\pm$ 342.1\\
        BootstrappedNN & 429.7 $\pm$ 162.2 & 179.1 $\pm$ 70.5   & 705.5  $\pm$ 409.8\\
        ParameterNoise & 467.5 $\pm$ 287.5 & 170.6 $\pm$ 65.4 & 785.51  $\pm$ 585.5\\
        \textbf{FunctionalPosterior} & \textbf{202.4 $\pm$ 70.0}&\textbf{ 98.46 $\pm$ 30.3} & 
        \textbf{252.8 $\pm$ 35.5}\\ 
        \bottomrule
    \end{tabular}
    \label{tab:regrets_summary}
    \end{center}
\end{table}

\vspace*{-10mm}
\section{Conclusion} 
In this work, we introduce a new deep Bayesian bandit approach for \textit{in-vitro} anticancer therapeutics selection. To enable efficient exploration in a contextual bandit setting with drug compound features and gene expression, we use functional variational approach to approximate this pharmacogenomics posterior. Empirically, we demonstrate that our approach explores efficiently and performs the best. The result is consistent across three publicly available pharmacogenomics datasets in cancer treatment recommendations. We hope that our work will serve as a stepstone towards improving the efficacy and quality of treatment allocation strategy in both pre-clinical therapeutic discovery and clinical study.




\newpage
\bibliography{sample}

\appendix
\section*{Appendix A.}



\section*{A.1 Contextual bandits in Healthcare }
An instantiation of the contextual bandits framework in treatment recommendations.
\begin{algorithm}[h!] 
\caption{ Contextual Bandits in Healthcare}
\begin{algorithmic}[1]
\State Available treatment options $a \in A$, and the treatment period length $T$.
\For{$t=1,2, \ldots, T$}
    \State Agent receives the context information $x_t$ for the current patient.
    \State A treatment $a_t$ is determined based on the history and the context information.
    \State Record the post-treatment health outcome $r_t$ for the patient.
\EndFor
\end{algorithmic}
\label{alg:banditsinhealthcare}
\end{algorithm}

\subsection*{A.2 Thompson Sampling \& Deep Bayesian Bandits} Deep Bayesian Bandits refers to posterior approximation with a deep neural network. For example, in Algorithm \ref{alg:thompsonSampling}, can be neural neural networks parameters drawn from a weight distribution, $P(\theta)$. \citet{riquelme2018deep} conduct a comprehensive analysis of different approximation methods in deep Bayesian bandit settings and propose \textit{NeuralLinear}, a Bayesian linear regression built on the representation of the last layer of a neural network. \citet {blundell2015weight} introduce \textit{Bayes by Backprop}, which learns the probability distribution of weights. \citet{DeepFPL} propose \emph{DeepFPL}, a neural network that uses ensemble sampling as an approximation to generating new noise in each round. In addition to posterior approximation of model weights, \citet{neuralts} propose Neural Thompson Sampling, which focuses on the posterior distribution of reward. \cite{sun2019functional} introduce the functional prior, sampling from the functional posterior by concatenating a random noise vector into input. 

\begin{algorithm}[h!]
\caption{Thompson Sampling}
\begin{algorithmic}[1]
\State $f(\cdot \mid \hat{\theta}_0); $ prior distribution over models, $P_0(\theta): \hat{\theta}_0 \in \Theta \rightarrow [0,1]$
\For{$t=1,2, \ldots, T$ }
    \State Observe context $x_t  \in R^{d}.$ 
    \State Sample model parameter $\hat{\theta}_t \sim P_{t}(\theta).$
    \State $a_t = \argmax_a f(x,a \mid \hat{\theta}_t)$. 
    \State Select action $a_t$ and observe reward $r_t(a_t)$. 
    \State Update the history $D_{t} \leftarrow D_{t-1} \cup (x_t, a_t, r_t(a_t)).$ 
    \State Update the posterior distribution $P_{t+1}(\theta) =\textbf{PosteriorUpdate}(D_{t},P_t).$
\EndFor
\end{algorithmic}
\label{alg:thompsonSampling}
\end{algorithm}

\section*{A.3 Dataset}\label{appen:dataset}
GDSC is the resource for therapeutic biomarker discovery in cancer cells. The GDSC1 dataset was generated between 2009 and 2015 using a matched set of cancer cell lines; contains 987 cell lines and 367 compounds. GDSC2 was generated in 2015 following improvements to screen design and assay \citep{IORIO2016740}; contains 198 Compounds and 809 cell lines. The Cancer Cell Line Encyclopedia (CCLE) project, started in 2008, contains pharmacological profiles for 24 anticancer drugs across 479 cell lines. For all datasets, we first filtered out cancer cell lines with missingness in drug-dose response greater than $70\%$. We then filtered out drug compounds with missing sensitivity in at least one of the curated cell lines. 
\begin{table}[h]
    \centering
    \begin{tabular}{lllll }
     Dataset & number of cell lines & number of drugs & Drug Response &\\
     \toprule
     GDSC1  & 472 & 133 & IC-50 &\\
     GDSC2  & 602 & 116 & IC-50 &\\
     CCLE   & 411 & 21  & Activity Area &\\
    \end{tabular}
    \caption{Summary Statistics of drug response dataset used in the work}
    \label{tab:my_label}
\end{table}

\section*{A.4 Baselines \& Benchmark Algorithms}\label{appen:baselines}

\paragraph{Uniform} a baseline for which an agent takes each action at random with equal probability.

\paragraph{NeuralGreedy} is a algorithm trained with a neural network and acts greedily (take action that predicted with the highest drug response.) where a random action is selected with probability $\epsilon$.

\paragraph{BayesByBackprop (BBB)} is proposed by \citet{blundell2015weight} and is one of the commonly used variational approaches that aproximate posterior by minimizing KL divergence. It is a Bayesian neural network with sampled weights from the variational posterior, $w\sim q(w|\theta)$. Variational parameters $\theta$ are updated with KL divergence between a mixture of two Gaussian densities prior and model weights distribution. BBB is a commonly used baseline in deep Bayesian bandits. 
\paragraph{Dropout} is a training technique where the output of each neuron is independently zeroed out with probability $p$ at each forward pass \citep{JMLR:v15:srivastava14a, Dropout}. Following the best action with respect to the random dropout prediction can be interpreted as an implicit form of Thompson sampling. 
\paragraph{BootstrappedNN}
is an empirical approach to approximate the posterior sampling distribution \citep{osband2015bootstrapped}. The idea is to train \textit{q} models with different dataset $D_i$ where $D_i$ is typically created by sampling with replacement from an original dataset, $D$. In our case, we train $q$ models with the same structure as \emph{NeuralGreedy} and $D$ is the history context that the agent has seen. In action selection, we sample a model uniformly at random (with probability $1/q$) and take action predicted reward to be the best by the sampled model.

\paragraph{DirectNoiseInjection} is a recently proposed method by \citet{plappert2018parameter}. Parameter-noise injection is an exploration technique proved successful in RL\cite{}. Through noise injection in parameter space: $\widetilde{\theta} = \theta + \calN(0,\sigma^2)$, we can perform state-dependent exploration. When selecting action, model weights are perturbed with isotropic Gaussian noise so that we get $a_t = \pi_{\Tilde{\theta}}(s_t)$. The magnitude of noise is adaptively adjusted based on distance measure between perturbed and non-perturbed policy in action space.\citep{plappert2018parameter}

\paragraph{ Network Structure and Hyper-Parameter Tuning}\label{oracle_policy}
For our oracle policy, we trained a 2-layer fully connected feedforward network with full access to genomics features, drug structure and corresponding drug response in the dataset. For each experiment, we ran 20 trials with 5,000 steps in each trials. All algorithms were updated every 30 steps. The size of each batch is 32. For all neural network structures, we adopted the same number of layers and hidden dimensions. as our oracle policy. We performed a grid search over parameters within a plausible range and reported the best results for each algorithm. We did a grid search over $\{ 1e^{-1}, 1e^{-2},1e^{-3}, 1e^{-4}, 1e^{-5}\}$ for learning rate. For NeuralGreedy, we performed grid search for $\epsilon$ over $\{ 0.05, 0.1, 0.2, 0.3, 0.4, 0.5, 0.6\}$. For DirectNoiseInjection, we searched noise range over $\{ 1e^{-1}, 1e^{-2},1e^{-3}, 1e^{-4}, 1e^{-5}\}$. For Dropout, we searched over probability $\{ 0.2, 0.4, 0.6, 0.8\}$. For BayesByBackprop, we searched prior noise over $\{ 1, 1e^{-1}, 1e^{-2}\}$ and prior mean over $\{ 1, 1e^{-1}, 1e^{-2}\}$. For BootstrappedNN, we searched the number of models over $\{ 2,3, 5, 10\}$.

\section*{A.5 Extra Results}

\begin{figure}[htp]
\centering
\includegraphics[width=.33\textwidth]{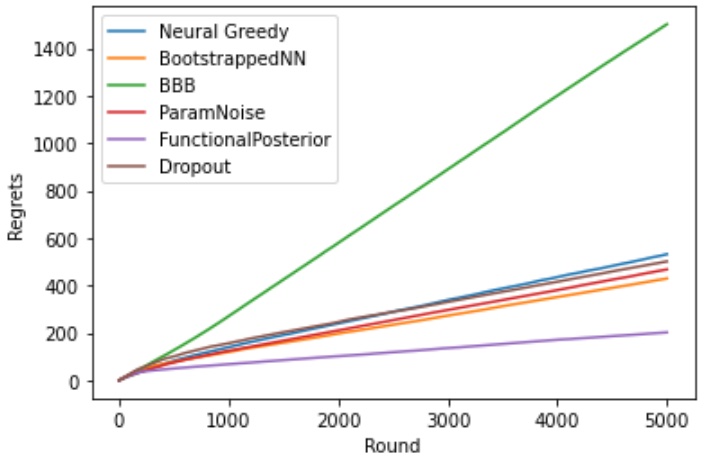}\hfill
\includegraphics[width=.33\textwidth]{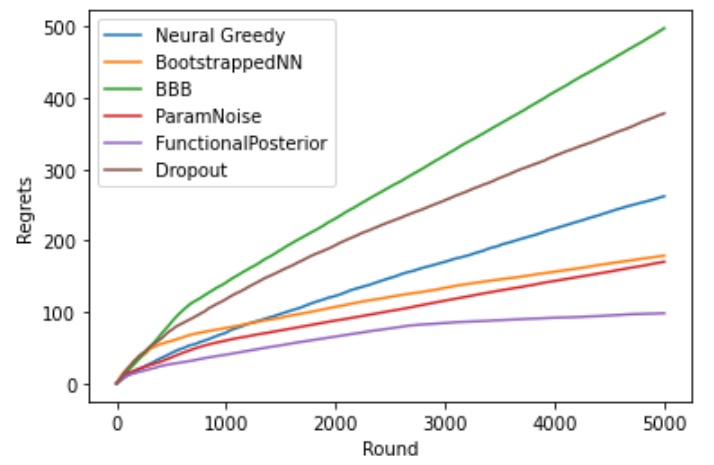}\hfill
\includegraphics[width=.33\textwidth]{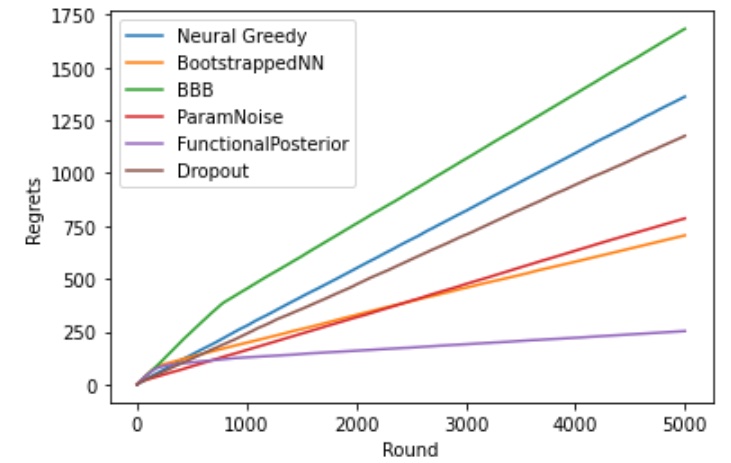}
\caption{Comparison of (mean) cumulative regrets at each step in maximizing cancer treatment effect for 5,000 rounds over 20 trials in GDSC-1, GDSC-2, \& CCLE. (left to right.)}
\label{fig:figure3}
\end{figure}

\end{document}